\documentclass[11pt,a4paper]{article}
\usepackage[hyperref]{acl2019}
\usepackage{times}
\usepackage{latexsym}

\usepackage{microtype}

\usepackage{graphicx}
\usepackage{bm}
\usepackage{url}
\usepackage{booktabs}
\usepackage{amsmath,amssymb,amsfonts}
\usepackage{xcolor}
\usepackage{tabularx}
\usepackage{arydshln}

\aclfinalcopy 
\def\aclpaperid{346} 

\title{Structural Information Preserving for Graph-to-Text Generation}

\author{Linfeng Song\textsuperscript{1}, Ante Wang\textsuperscript{2}, Jinsong Su\textsuperscript{2}\thanks{~~Corresponding author}, Yue Zhang\textsuperscript{3}, Kun Xu\textsuperscript{1}, Yubin Ge\textsuperscript{4} and Dong Yu\textsuperscript{1} \\
  \textsuperscript{1}Tencent AI Lab, Bellevue, WA, USA \\
  \textsuperscript{2}Xiamen University, Xiamen, China \\
  \textsuperscript{3}School of Engineering, Westlake University, China \\
  \textsuperscript{4}University of Illinois at Urbana-Champaign, USA \\}

\date{}

\begin{document}
\maketitle
\begin{abstract}
  The task of graph-to-text generation aims at producing sentences that preserve the meaning of input graphs.
  As a crucial defect, the current state-of-the-art models may mess up or even drop the core structural information of input graphs when generating outputs.
  We propose to tackle this problem by leveraging richer training signals that can guide our model for preserving input information.
  In particular, we introduce two types of autoencoding losses, each individually focusing on different aspects (a.k.a. views) of input graphs.
  The losses are then back-propagated to better calibrate our model via multi-task training.
  Experiments on two benchmarks for graph-to-text generation show the effectiveness of our approach over a state-of-the-art baseline.
  Our code is available at \url{http://github.com/Soistesimmer/AMR-multiview}.
\end{abstract}

\section{Introduction}

\begin{figure}
    \centering
    \includegraphics[width=\linewidth]{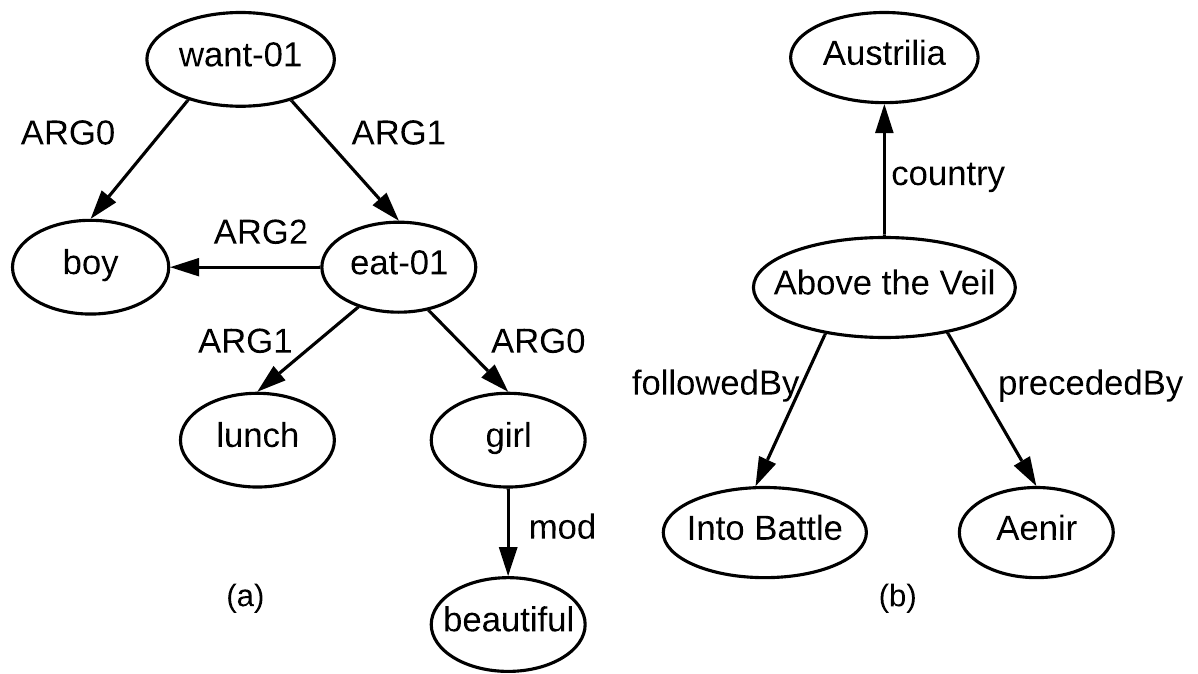}
    \caption{(a) An AMR graph meaning ``\emph{The boy wants the beautiful girl to eat lunch with him.}'', and (b) A knowledge graph carrying the meaning ``\emph{Above the Veil is an Australian novel and the sequel to Aenir. It was followed by Into the Battle.}''}
    \label{fig:example}
    \vspace{-1.0em}
\end{figure}

Many text generation tasks take graph structures as their inputs, such as Abstract Meaning Representation (AMR) \citep{banarescu2013abstract}, Knowledge Graph (KG) and database tables.
For example, as shown in Figure \ref{fig:example}(a), AMR-to-text generation is to generate a sentence that preserves the meaning of an input AMR graph, which is composed by a set of concepts (such as ``\emph{boy}'' and ``\emph{want-01}'') and their relations (such as ``\emph{:ARG0}'' and ``\emph{:ARG1}'').
Similarly, as shown in Figure \ref{fig:example}(b), KG-to-text generation is to produce a sentence representing a KG, which contains worldwide factoid information of entities (such as ``\emph{Australia}'' and ``\emph{Above the Veil}'') and their relations (such as ``\emph{followedBy}'').

Recent efforts on graph-to-text generation tasks mainly focus on how to effectively represent input graphs, so that an attention mechanism can better transfer input knowledge to the decoder when generating sentences.
Taking AMR-to-text generation as an example, different graph neural networks (GNNs) \citep{beck2018graph,song2018graph,guo2019densely,ribeiro2019enhancing} have been introduced to better represent input AMRs than a sequence-to-sequence model \citep{konstas2017neural}, and later work \citep{zhu2019modeling,cai2019graph,wang2020amr} showed that relation-aware Transformers can achieve even better results than GNNs.
These advances for encoding have largely pushed the state-of-the-art performance.

Existing models are optimized by maximizing the conditional word probabilities of a reference sentence, a common signal for training language models.
As a result, these models can learn to produce fluent sentences, but some crucial input concepts and relations may be messed up or even dropped.
Taking the AMR in Figure \ref{fig:example}(a) as an example, a model may produce ``\emph{the girl wants the boy to go}'', which conveys an opposite meaning to the AMR graph.
In particular, this can be very likely if ``\emph{the girl wants}'' appears much more frequent than ``\emph{the boy wants}'' in the training corpus.
This is a very important issue, because of its wide existence across many neural graph-to-text generation models, hindering the usability of these models for real-world applications \citep{duvsek2018findings,duvsek2019semantic,balakrishnan-etal-2019-constrained}.

A potential solution for this issue is improving the training signal to enhance preserving of structural information.
However, little work has been done to explore this direction so far, probably because designing such signals is non-trivial.
As a first step towards this goal, we propose to enrich the training signal with additional \emph{autoencoding losses} \citep{rei2017semi}.
Standard autoencoding 
for graph-to-sequence tasks 
requires reconstructing (parsing into) input graphs, while the parsing algorithm for one type of graphs (such as knowledge graphs) may not generalize to other graph types or may not even exist.
To make our approach general across different types of graphs, we propose to reconstruct different \emph{views} of each input graph (rather than the original graph), where each view highlights one aspect of the graph and is easy to produce.
Then through multi-task learning, the autoencoding losses of all views are back-propagated to the whole model so that the model can better follow the input semantic constraints.

Specifically, we break each input graph into a set of triples to form our first view, where each triple (such as ``\emph{want-01 :ARG0 boy}'' in Figure \ref{fig:example}(a)) contains a pair of entities and their relation.
As the next step, the alignments between graph nodes and target words are generated to ground this view into the target sentence for reconstruction.
Our second view is the linearization of each input graph produced by depth-first graph traversal, and this view is reconstructed token-by-token from the last decoder state.
Overall the first view highlights the \emph{local} information of each triple relation, the second view focuses on the \emph{global} semantic information of the entire graph.

Experiments on AMR-to-text generation and WebNLG \citep{gardent2017webnlg} show that our graph-based multi-view autoencoding loss improves the performance of a 
state-of-the-art baseline by more than 2 BLEU points
without introducing any parameter during decoding.
Besides, human studies show that our approach is indeed beneficial for preserving more concepts and relations from input graphs.

\section{Related Work}

Previous work for neural graph-to-text generation \citep{konstas2017neural,song2018graph,beck2018graph,trisedya2018gtr,marcheggiani2018deep,xu2018sql,cao2019factorising,damonte2019structural,hajdik2019neural,koncel2019text,hong2019improving,song2019semantic,su2017lattice} mainly studied how to effectively represent input graphs, and all these models are trained only with the standard language modeling loss.
As the most similar one to our work, \citet{tu2017neural} proposed an encoder-decoder-reconstructor model for machine translation, which is trained not only to translate each source sentence into its target reference, but also to translate the target reference back into the source text (reconstruction).
\citet{wiseman2017challenges} extended 
the reconstruction loss of 
\citet{tu2017neural} on table-to-text generation, where a table contains multiple records that fit into several fields.
We study a more challenging topic on how to reconstruct a complex graph structure rather than a sentence or a table, and we propose two general and effective methods that reconstruct different complementary views of each input graph.
Besides, we propose methods to breakdown the whole (graph, sentence) pair into smaller pieces of (edge, word) pairs with alignments, before training our model to reconstruct each edge given the corresponding word.
On the other hand, neither of the previous efforts tried to leverage this valuable information.

Our work is remotely related to the previous efforts on string-to-tree neural machine translation (NMT) \citep{aharoni2017towards,wu2017sequence,wang2018tree}, which aims at generating target sentences with their syntactic trees.
One major difference is that their goal is producing grammatical outputs, while ours is preserving input structures.
Besides, our multi-view reconstruction framework is a \emph{detachable} component on top of the decoder states for training, so no extra error propagation (for structure prediction) can be introduced.
Conversely, their models generate trees together with target sentences, thus extra efforts \citep{wu2017sequence} are introduced to alleviate error propagation.
Finally, there exist transition-based algorithms \citep{nivre2003efficient} to convert tree parsing into the prediction of transition actions, while
we study reconstructing graphs, where there is no common parsing algorithm for all graph types.

Autoencoding loss by input reconstruction was mainly adopted on sequence labeling tasks, such as named entity recognition (NER) \citep{rei2017semi,liu2018empower,jia2019cross}, simile detection \citep{liu2018neural} and sentiment analysis \citep{rei2019jointly}.
Since input reconstruction is not intuitively related to these tasks, the autoencoding loss only serves as more training signals.
Different from these efforts, we leverage autoencoding loss as a means to preserve input knowledge.
Besides, we study reconstructing complex graphs, proposing a general multi-view approach for this goal.


\begin{figure*}
    \centering
    \includegraphics[width=\textwidth]{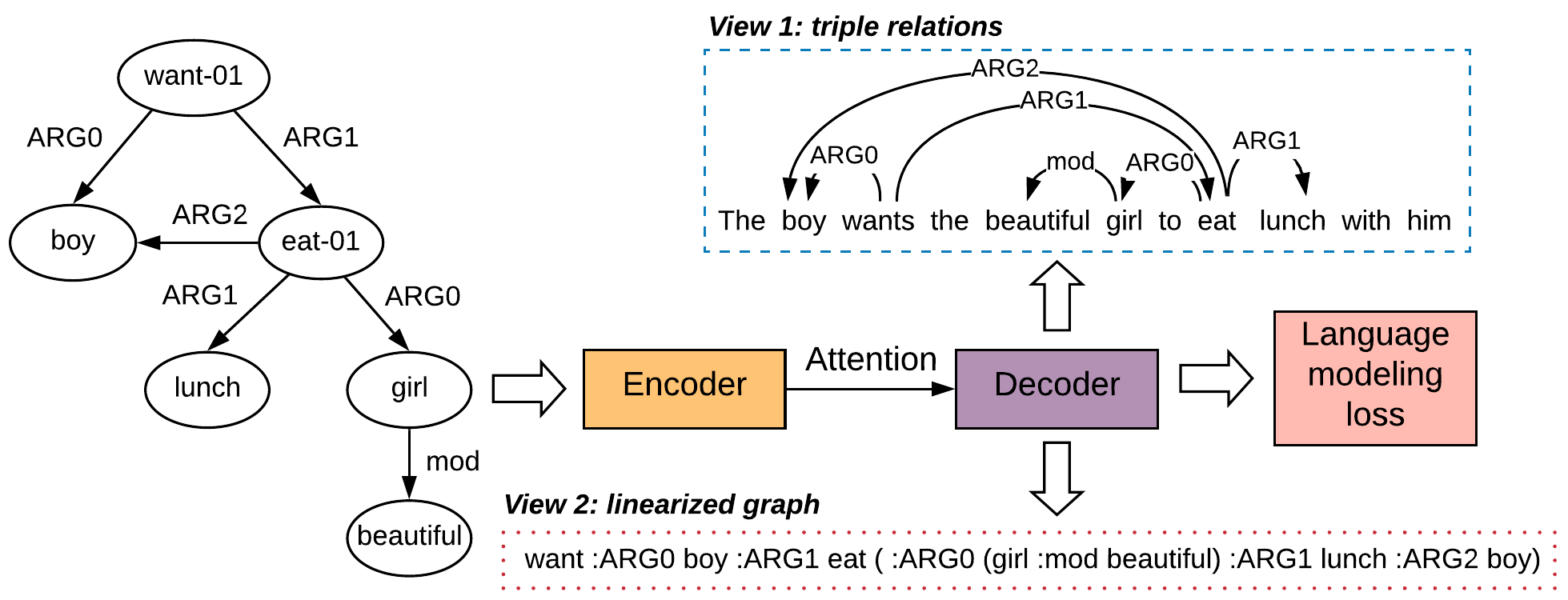}
    \caption{The training framework using multi-view autoencoding losses.}
    \label{fig:model}
\end{figure*}

\section{Base: Structure-Aware Transformer}
\label{sec:baseline}

Formally, an input for graph-to-text generation can be represented as $\boldsymbol{G}=\langle \boldsymbol{V}, \boldsymbol{E}\rangle$, where $\boldsymbol{V}$ is the set of graph nodes and $\boldsymbol{E}$ corresponds to all graph edges.
Each edge $e\in\boldsymbol{E}$ is a triple $(v_i,l,v_j)$, showing labelled relation between two connected nodes $v_i$ and $v_j$.
Given a graph, we choose a recent relation-aware transformer model \citep{zhu2019modeling} as our baseline to generate the ground-truth sentence $\boldsymbol{y}=(y_1,\dots,y_N)$ that contain the same meaning as the input graph.
It exhibits the state-of-the-art performance for AMR-to-text generation.

\subsection{Structure-aware Transformer Encoder}

Similar to the standard model
\citep{vaswani2017attention}, the structure-aware Transformer encoder stacks multiple self-attention layers on top of an embedding layer to encode linearized graph nodes.
Taking the $l$-th layer for example, it consumes the states of its preceding layer ($\boldsymbol{h}_1^{l-1}\dots \boldsymbol{h}_N^{l-1}$, or the embedding layer when $l$ is 1) and its states are then updated by a weighted sum:
\begin{equation}
    \boldsymbol{h}_i^{l} = \sum_{j\in[1..N]} \alpha_{ij}\large( \boldsymbol{h}_j^{l-1} \boldsymbol{W}^P + \boldsymbol{\gamma}_{ij}\boldsymbol{W}^{R_1} \large) \text{,}
\end{equation}
where $\boldsymbol{\gamma}_{ij}$ is the vector representation of the relation between nodes $v_i$ and $v_j$, and $\boldsymbol{W}^P$ and $\boldsymbol{W}^{R_1}$ are model parameters.
The weights, such as $\alpha_{ij}$, are obtained by relation-aware self-attention:
\begin{align}
    \alpha_{ij} &= \frac{\exp(e_{ij})}{\sum_{k\in[1..N]}\exp(e_{ik})} \\
    e_{ij} &= \frac{\big(\boldsymbol{h}_i^{l-1}\boldsymbol{W}^Q\big)\big(\boldsymbol{h}_j^{l-1}\boldsymbol{W}^K+\boldsymbol{\gamma}_{ij}\boldsymbol{W}^{R_2}\big)^\intercal}{\sqrt{d_h}} 
\end{align}
where $\boldsymbol{W}^Q$, $\boldsymbol{W}^K$ and $\boldsymbol{W}^{R_2}$ are model parameters, and $d_h$ denotes the encoder-state dimension.
The encoder adopts $L$ self-attention layers and $\boldsymbol{H}^L=$ $(\boldsymbol{h}_1^{L}\dots \boldsymbol{h}_{|\boldsymbol{V}|}^{L})$ represents the concatenated top-layer hidden states of the encoder, which will be used in attention-based decoding.

Compared with the standard model, this encoder introduces the vectorized structural information (such as $\boldsymbol{\gamma}_{ij}$) for all node pairs.
Given a node pair $v_i$ and $v_j$, generating such information involves two main steps.
First, a sequence of graph edge labels along the path from $v_i$ to $v_j$ are obtained, where a direction symbol is added to each label to distinguish the edge direction.
For instance, the label sequence from ``\emph{boy}'' to ``\emph{girl}'' in Figure \ref{fig:example}(a) is ``\emph{:ARG0$\uparrow$ :ARG1$\downarrow$ :ARG0$\downarrow$}''.
As the next step, the label sequence is treated as a single (feature) token and represented by the corresponding embedding vector, and this vector is taken as the vectorized structural information $\boldsymbol{\gamma}_{ij}$ from $v_i$ to $v_j$.
Since there are a large number of features, only the most frequent 20K are kept, while the rest are mapped into a special \emph{UNK} feature.\footnote{\citet{zhu2019modeling} also mentions other (such as CNN-based or self-attention-based) alternatives to calculate $\boldsymbol{\gamma}_{ij}$. While the GPU memory consumption of these alternatives is a few times more than our baseline, ours actually shows a comparable performance.}

\subsection{Standard Transformer Decoder}

The decoder is the same as the standard Transformer architecture, which stacks an embedding layer, multiple ($L$) self-attention layers and a linear layer with softmax activation to generate target sentences in a word-by-word manner.
Each target word $y_i$ and decoder state $\boldsymbol{s}_i$ are generated sequentially by the self-attention decoder:
\begin{equation} \label{eq:target_decoder}
    y_i, \boldsymbol{s}_i = \mathtt{SADecoder}([\boldsymbol{H}^{L};\boldsymbol{s}_1...\boldsymbol{s}_{i-1}], y_{i-1}) \text{,}
\end{equation}
where $\mathtt{SADecoder()}$ is the function of decoding one step with the self-attention-based decoder.

\subsection{Training with Language Modeling Loss}

Same as most previous work, this model is trained with the standard language modeling loss that minimizes the negative log-likelihood of conditional word probabilities:
\begin{equation} \label{eq:lm_loss}
\begin{split}
    l_{base} &= -\sum_{i\in[1..N]} \log p(y_i|y_1,...,y_{i-1};\boldsymbol{G}) \\
    &= -\sum_{i\in[1..N]} p(y_i|\boldsymbol{s}_i;\boldsymbol{\theta}) \text{,}
\end{split}
\end{equation}
where $\boldsymbol{\theta}$ represents all model parameters.

\section{Multi-View Autoencoding Losses}

Figure \ref{fig:model} visualizes the training framework using our multi-view autoencoding losses, where the attention-based encoder-decoder model with the language modeling loss is the baseline.
Our losses are produced by reconstructing the two proposed views (surrounded by slashed or dotted box) of the input graph, where each view represents a different aspect of the input.
With the proposed losses, we expect to better refine our model for preserving the structural information of input graphs.

\subsection{Loss 1: Reconstructing Triple Relations with Biaffine Attention}

Our first view breaks each input graph into a set of triples, where each triple (such as ``\emph{want-01 :ARG0 boy}'' in Figure \ref{fig:example}(a)) contains a pair of nodes and their labeled relation.
Next, we use pre-generated alignments between graph nodes and target words to ground the graph triples onto the target sentence.
As illustrated in the slashed blue box of Figure \ref{fig:model}, the result contains several labeled arcs, each connecting a word pair (such as ``\emph{wants}'' and ``\emph{boy}'').
While each arc represents a \emph{local} relation, their combination implies the \emph{global} input structure.

For certain types of graphs, a node can have multiple words.
To deal with this situation, we use the first word of both associated graph nodes when grounding a graph edge onto the target sentence.
Next, 
we also connect the first word of each grounded entity with the other words of the entity in order to represent the whole-entity information in the sentence.
Taking the edge ``\emph{followedBy}'' in Figure \ref{fig:example}(b) as an example, we first ground it onto the target sentence to connect words ``\emph{Above}'' and ``\emph{Into}''.
Next, we create edges with label ``\emph{compound}'' from ``\emph{Above}'' to words ``\emph{the}'' and ``\emph{Veil}'', and from ``\emph{Into}'' to words ``\emph{the}'' and  ``\emph{Battle}'' to indicate the two associated entity mentions.

For many tasks on graph-to-text generation, the node-to-word alignments can be easily generated from off-the-shell toolkits.
For example, in AMR-to-text generation, there have been several aligners \citep{pourdamghani2014aligning,flanigan2016cmu,wang2017getting,liu2018amr,szubert2018structured} available for linking AMR nodes to words.
For knowledge graphs, the alignments can be produced by simple rule-based matching or an entity-linking system.

The resulting structure with labeled arcs connecting word pairs resembles a dependency tree, and thus we employ a deep biaffine model \citep{dozat2016deep} to predict this structure from the decoder states.
More specifically, the model factorizes the probability for making each arc into two parts: an unlabeled factor and a labeled one.
Given the decoder states $\boldsymbol{s}_1,\dots,\boldsymbol{s}_N$, the representation for each word $y_i$ as the head or the modifier of any unlabeled factor is calculated by passing its hidden state $\boldsymbol{s}_i$ through the corresponding multi-layer perceptrons (MLPs):
\begin{align} \label{eq:arc-h}
    \boldsymbol{r}_i^{\mathrm{arc-h}} &= \mathtt{MLP}^{\mathrm{arc-head}}(\boldsymbol{s}_i) \\ \label{eq:arc-m}
    \boldsymbol{r}_i^{\mathrm{arc-m}} &= \mathtt{MLP}^{\mathrm{arc-mod}}(\boldsymbol{s}_i) \text{,}
\end{align}
The (unnormalized) scores for the unlabeled factors with any possible head
word given the modifier $y_i$ are calculated as:
\begin{equation}
        \boldsymbol{\phi}_i^{\mathrm{arc}} = \boldsymbol{R}^{\mathrm{arc-h}}\boldsymbol{U}_a^\intercal \boldsymbol{r}_i^{\mathrm{arc-m}} + \boldsymbol{R}^{\mathrm{arc-h}}\boldsymbol{v}_a \text{,}
\end{equation}
where $\boldsymbol{R}^{\mathrm{arc-h}}$ is the concatenation of all $\boldsymbol{r}_i^{\mathrm{arc-h}}$, and $\boldsymbol{U}_a$ and $\boldsymbol{v}_a$ are model parameters.
Similarly, the representations for word $y_i$ being the head or the modifier of a labeled factor are calculated by two additional MLPs:
\begin{align} \label{eq:label-h}
    \boldsymbol{r}_i^{\mathrm{label-h}} &= \mathtt{MLP}^{\mathrm{label-head}}(\boldsymbol{s}_i) \\ \label{eq:label-m}
    \boldsymbol{r}_i^{\mathrm{label-m}} &= \mathtt{MLP}^{\mathrm{label-mod}}(\boldsymbol{s}_i) \text{,}
\end{align}
and the (unnormalized) scores for all relation labels given the head word $y_j$
and the modifier $y_i$ are calculated as:
\begin{multline}
    \boldsymbol{\phi}_{i,j}^{\mathrm{label}} = \boldsymbol{r}_j^{\mathrm{label-h}} \boldsymbol{U}_l \boldsymbol{r}_i^{\mathrm{label-m}} + \\ (\boldsymbol{r}_j^{\mathrm{label-h}} \oplus \boldsymbol{r}_i^{\mathrm{label-m}})^\intercal \boldsymbol{V}_l + \boldsymbol{b}_l,
\end{multline}
where $\boldsymbol{U}_l$, $\boldsymbol{V}_l$ and $\boldsymbol{b}_l$ are model parameters.
The overall conditional probability of a labeled arc with label $l$, head word $y_j$ and modifier $y_i$ is calculated by the following chain rule:
\begin{multline}
    p(y_j,l|y_i) = p(l|y_j,y_i) \cdot p(y_j|y_i) \\
    = \mathtt{softmax}(\boldsymbol{\phi}_{i,j}^{\mathrm{label}})_{[l]} \cdot \mathtt{softmax}(\boldsymbol{\phi}_i^{\mathrm{arc}})_{[j]},
\end{multline}
where $[x]$ in the subscript represents choosing the $x$-th item from the corresponding vector.

As the final step, the loss for reconstructing this view is defined as the negative log-likelihood of all target arcs $\boldsymbol{E}^\prime$ (the grounded triples from $\boldsymbol{E}$):
\begin{equation}
    l_{auto1} = \sum_{(y_j,l,y_i)\in \boldsymbol{E}^\prime} -\log p(y_j,l|y_i)
\end{equation}

\subsection{Loss 2: Reconstructing Linearized Graphs with a Transformer Decoder}

As a supplement to our first loss for reconstructing the \emph{local} information of each grounded triple, we introduce the second loss for predicting the whole graph as a linearized sequence.
To minimize the loss of the graph structural information caused by linearization, we adopt an algorithm based on depth-first traversal \citep{konstas2017neural}, which inserts brackets to preserve graph scopes.
One linearized AMR graph is shown in the red dotted box of Figure \ref{fig:model}, where the node suffixes (such as ``-01'') representing word senses are removed.

One may argue that we could directly predict the original graph so that no structural information would be lost.
However, each type of graphs can have their own parsing algorithm due to their unique properties (such as directed vs undirected, rooted vs unrooted, etc).
Such an exact prediction will hurt the generality of the proposed approach.
Conversely, our solution is general, as linearization works for most types of graphs.
From Figure \ref{fig:model} we can observe that the inserted brackets clearly infer the original graph structure.
Besides, previous work \citep{iyer2017learning,konstas2017neural} has shown the effectiveness of generating linearized graphs as sequences for
graph parsing, which also confirms our observation.

Given a linearized graph represented as a sequence of tokens $x_1,\dots,x_M$, 
where each token $x_i$ can be a graph node, a edge label or a inserted bracket, 
we adopt another standard Transformer decoder ($\mathtt{SADecoder}_{g}$) to produce the sequence:
\begin{equation} \label{eq:linearize_decoder}
    x_i, \boldsymbol{t}_i = \mathtt{SADecoder}_{g}([\boldsymbol{S};\boldsymbol{t}_1...\boldsymbol{t}_{i-1}],x_{i-1}) \text{,}
\end{equation}
where $\boldsymbol{S}=(\boldsymbol{s}_1\dots\boldsymbol{s}_N)$ denotes the concatenated states for the target sentence (Equation \ref{eq:target_decoder}), and the loss for reconstructing this view is defined as the negative log-likelihood for the linearized graph:
\begin{equation}
    l_{auto2} = - \sum_{i\in[1..M]} \log p(x_i|\boldsymbol{t}_i;\boldsymbol{\theta}) \text{,}
\end{equation}
where $\boldsymbol{\theta}$ represents model parameters.

\subsection{Discussion and Comparison}

Our autoencoding modules function as \emph{detachable} components based on the target-side decoder states, and thus this brings two main benefits.
First, our approaches are not only orthogonal to the recent advances \citep{li2015gated,kipf2016semi,velivckovic2017graph} on the encoder side for representing graphs, but also flexible with other decoders based on multi-layer LSTM \citep{hochreiter1997long} or GRU \citep{cho2014learning}.
Second, no extra error propagation is introduced, as our approach does not affect the normal sentence-decoding process.

In addition to the different aspects both losses focus on, each has some merits and disadvantages over the other.
In terms of training speed, calculating \emph{Loss 1} can be faster than \emph{Loss 2}, because predicting the triple relations can be done in parallel, while it is not feasible for generating a linearized graph.
Besides, calculating \emph{Loss 1} suffers from less variances, as the triple relations are agnostic to the token order determined by input files.
Conversely, graph linearization is highly sensitive to the input order.
One major merit for \emph{Loss 2} is the generality, as node-to-word alignments may not be easily obtained, especially for multi-lingual tasks.

\subsection{Training with Autoencoding Losses}

The final training signal with both proposed autoencoding losses is formalized as:
\begin{equation}
    l_{final} = l_{base} + \alpha l_{auto1} + \beta l_{auto2} \text{,}
\end{equation}
where $\alpha$ and $\beta$ are coefficients for our proposed losses.
Both coefficient values are selected by a development experiment.

\section{Experiments}

We study the effectiveness of our autoencoding training framework on AMR-to-text generation and KG-to-text generation.
BLEU \cite{papineni2002bleu} and Meteor \cite{denkowski:lavie:meteor-wmt:2014} scores are reported for comparison. Following previous work, we use the \texttt{multi-bleu.perl} from Moses\footnote{http://www.statmt.org/moses/} for BLEU evaluation.

\subsection{Data}

\subparagraph{AMR datasets\footnote{https://amr.isi.edu/download.html}} 
We take LDC2015E86 that contains 16,833, 1,368 and 1,371 instances for training, development and testing, respectively.
Each instance contains a sentence and an AMR graph.
Following previous work, we use a standard AMR simplifier \citep{konstas2017neural} to preprocess our AMR graphs, and take the PTB-based Stanford tokenizer\footnote{https://nlp.stanford.edu/software/tokenizer.shtml} to tokenize the sentences.
The node-to-word alignments are produced by the ISI aligner \citep{pourdamghani2014aligning}.
We use this dataset for our \emph{primary} experiments.
We also report our numbers on LDC2017T10, a later version of AMR dataset that has 36521, 1,368 and 1,371 instances for training, development and testing, respectively.

\subparagraph{WebNLG \citep{gardent2017webnlg}}
This dataset consists of 18,102 training and 871 development KG-text pairs, where each KG is a subgraph of DBpedia\footnote{https://wiki.dbpedia.org/} that can contain up to 7 relations (triples).
The testset has two parts: \emph{seen}, containing 971 pairs where the KG entities and relations belong to the DBpedia categories that are seen in the training data, and \emph{unseen}, where the entities and relations come from unseen categories.
Same as most previous work, we evaluate our model on the \emph{seen} part, and this is also more relevant to our setup.

We follow \citet{marcheggiani2018deep} to preprocess the data. 
To obtain 
the alignments between a KG and a sentence,
we use a method based on heuristic string matching.
For more detail, we remove any abbreviations from a KG node (such as ``\emph{New York (NY)}'' is changed to ``\emph{New York}''), before finding the first phrase in the sentence that matches the longest prefix of the node.
As a result, we find a match for 91\% KG nodes.

\subsection{Settings}

For model hyperparameters, we follow the setting of our baseline \citep{zhu2019modeling}, where 6 self-attention layers are adopted with 8 heads for each layer.
Both sizes of embedding and hidden states are set to 512, and the batch token-size is 4096.
The embeddings are randomly initialized and updated during training.
All models are trained for 300K steps using Adam \citep{kingma2014adam} with $\beta1=0.1$.
Byte-pair encoding (BPE) \citep{sennrich2016neural} with 10K operations is applied to all datasets.
We use 1080Ti GPUs for experiments.

\begin{figure}
    \centering
    \includegraphics[width=\linewidth]{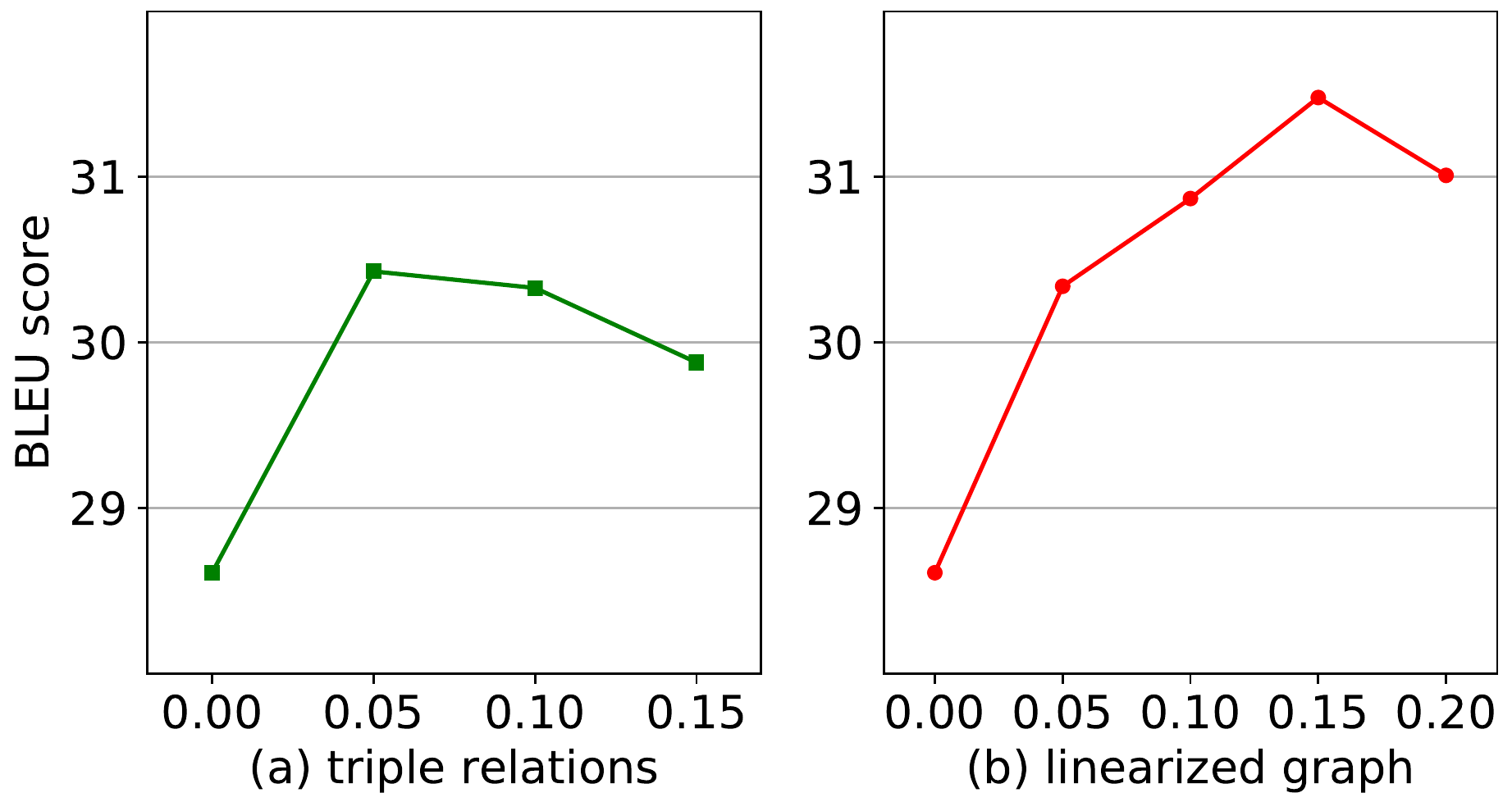}
    \caption{Development results on LDC2015E86.}
    \label{fig:dev_results}
\end{figure}

For our approach, the multi-layer perceptrons for deep biaffine classifiers (Equations \ref{eq:arc-h}, \ref{eq:arc-m}, \ref{eq:label-h} and \ref{eq:label-m}) take two layers of 512 units.
The Transformer decoder (Equation \ref{eq:linearize_decoder}) for predicting linearized graphs takes the same embedding and hidden sizes as the baseline decoder (Equation \ref{eq:target_decoder}).

\subsection{Development Results}

Figure \ref{fig:dev_results} shows the devset performances of using either \emph{Loss 1} (triple relations) or \emph{Loss 2} (linearized graph) under different coefficients.
It shows the baseline performance when a coefficient equals to 0.
There are large improvements in terms of BLEU score when increasing the coefficient of either loss from 0.
These results indicate the effectiveness of our autoencoding training framework.
The performance of our model with either loss slightly goes down when further increasing the coefficient.
One underlying reason is that an over-large coefficient will dilute the primary signal on language modeling, which is more relevant to the BLEU metric.
Particularly, we observe the highest performances when $\alpha$ and $\beta$ are 0.05 and 0.15, respectively, and thus we set our coefficients using these values for the remaining experiments.

\subsection{Main Results}

\begin{table}
\setlength\tabcolsep{4pt}
    \centering
    \begin{tabular}{lcc}
    \toprule
    Model & BLEU & Time \\
    \midrule
    LSTM \citep{konstas2017neural} & 22.00 & -- \\
    GRN \citep{song2018graph} & 23.28 & -- \\
    DCGCN \citep{guo2019densely} & 25.70 & -- \\
    RA-Trans-SA \citep{zhu2019modeling} & 29.66 & -- \\
    \hline
    RA-Trans-F-ours & 29.11 & 0.25 \\
    ~~~~~+~Loss 1 (triple relations) & 30.47 & 0.38 \\
    ~~~~~+~Loss 2 (linearized graph) & 31.13 & 0.52 \\
    ~~~~~+~Both & \textbf{31.41} & 0.61 \\
    \hline
    DCGCN (0.3M) & 33.2 & -- \\
    GRN (2M) & 33.6 & -- \\
    LSTM (20M) & 33.8 & -- \\
    DCGCN (ensemble, 0.3M) & 35.3 & -- \\
    \bottomrule
    \end{tabular}
    \caption{Main test results on LDC2015E86. Numbers such as ``\emph{2M}'' means the number of extra silver data being used, and ``\emph{ensemble}'' indicates model ensemble.}
    \label{tab:main}
    \vspace{-0.5em}
\end{table}

Table \ref{tab:main} shows the main comparison results with existing work for AMR-to-text generation, where ``\emph{Time}'' represents the average time (seconds) for training one step.
The first group corresponds to the reported numbers of previous models on this dataset, and their main difference is the encoder for presenting graphs:
\emph{LSTM} \citep{konstas2017neural} applies a multi-layer LSTM on linearized AMRs, \emph{GRN} \citep{song2018graph} and \emph{DCGCN} \citep{guo2019densely} adopt graph neural networks to encode original AMRs, and \emph{RA-Trans-SA} is the best performing model of \citet{zhu2019modeling}, 
using self attention to model the relation path for each node pair.

The second group reports our systems, where the \emph{RA-Trans-F-ours} baseline is our implementation of the feature-based model of \citet{zhu2019modeling}.
It shows a highly competitive performance on this dataset.
Applying \emph{Loss 1} alone achieves an improvement of 1.36 BLEU points, and \emph{Loss 2} alone obtains 0.66 more points than \emph{Loss 1}.
One possible reason is that \emph{Loss 2}, which aims to reconstruct the whole linearized graph, can provide more informative features.
Using both losses, we observe roughly a 2.3-point gain in terms of BLEU, indicating that both losses are complementary.

Regarding Meteor, \emph{RA-Trans-SA} reports 35.45, the highest among all previously reported numbers.
The \emph{RA-Trans-F-ours} baseline gets 35.0 that is slightly worse than \emph{RA-Trans-SA}.
Applying \emph{Loss 1} or \emph{Loss 2} alone gives a number of 35.5 and 36.1, respectively.
Using both losses, our approach achieves 36.2 that is better than \emph{RA-Trans-SA}.

Regarding the training speed, adopting \emph{Loss 2} requires double amount of time compared with the baseline, being much slower than \emph{Loss 1}.
This is because the biaffine attention calculations for different word pairs are parallelizable, while it is not for producing a linearized graph.
Using both losses together, we observe a moderately longer training process (1.4-times slower) than the baseline.
Please note that our autoencoding framework only affects the offline training procedure, leaving the online inference process unchanged.

The last group shows additional higher numbers produced by systems that use the ensemble of multiple models and/or additional silver data.
They suffer from problems such as requiring massive computation resources and taking a long time for training.
We leave exploring additional silver data and ensemble for further work.

\subsection{Quantitative Human Study on Preserving Input Relation}

\begin{table}
    \centering
    \begin{tabular}{lc}
    \toprule
    Model & Recall (\%) \\
    \midrule
    RA-Trans-F-ours & 78.00 \\
    ~~~~~+~Both & \textbf{85.13} \\
    \bottomrule
    \end{tabular}
    \caption{Human study for the recall of input relations on LDC2015E86.}
    \label{tab:human_study}
    \vspace{-1.0em}
\end{table}

Our multi-view autoencoding framework aims at preserving input relations, thus we further conduct a quantitative human study to estimate this aspect.
To this end, we first extract all interactions of a subject, a predict and an object (corresponding to the AMR fragment ``\emph{pred :ARG0 subj :ARG1 obj}'') from each AMR graph, and then check how many interactions are preserved by the output of a model.
The reason for considering this type of interaction comes from two folds: first, they convey fundamental information forming the backbone of a sentence, and second, 
they can be easily extracted from graphs and evaluated by human judges.

As shown in Table \ref{tab:human_study}, we choose 200 AMR-sentence pairs to conduct this study and compare our model with the baseline in terms of the recall number, showing the percent of preserved interactions.
To determine if a sentence preserves an interaction, we ask 3 people with NLP background to make their decisions and choose the majority vote as the human judgement.
Out of the 491 interactions, the baseline only preserves 78\%.
With our multi-view autoencoding losses, 7.13\% more interactions are preserved, which further confirms the effectiveness of our approach.

\subsection{Case Study}

As shown in Table \ref{tab:case_study}, we further demonstrate several typical examples from our human study for better understanding how our framework helps preserve structural input information.
Each example includes an input AMR, a reference sentence (\textbf{Ref}), the baseline output (\textbf{Baseline}) and the generated sentence by our approach (\textbf{Our approach}).

\begin{table}[t!] \fontsize{10}{11} \selectfont
    \centering
    \begin{tabularx}{0.96\linewidth}{X}
     \toprule
     (r / recommend-01 \\
     ~~~~:ARG0 (i / i) \\
     ~~~~:ARG1 (g / go-02 \\
     ~~~~~~~~:ARG0 (y / you) \\
     ~~~~~~~~:purpose (s / see-01 \\
     ~~~~~~~~~~~~:ARG0 y \\
     ~~~~~~~~~~~~:ARG1 (p / person \\
     ~~~~~~~~~~~~~~~~:ARG0-of (h / have-rel-role-91 \\
     ~~~~~~~~~~~~~~~~~~~~:ARG1 y \\
     ~~~~~~~~~~~~~~~~~~~~:ARG2 (d / doctor))) \\
     ~~~~~~~~~~~~:mod (t / too))) \\
     ~~~~:ARG2 y) \\
     \textbf{Ref}: i 'd recommend you go and see your doctor too . \\
     \textbf{Baseline}: \textcolor{red}{i should go to see} your doctor too . \\
     \textbf{Our approach}: \textcolor{blue}{i recommend you to go to see} your doctor too . \\
     \midrule
     (c / country \\
     ~~~~:mod (o / only) \\
     ~~~~:ARG0-of (h / have-03 \\
     ~~~~~~~~:ARG1 (p / policy \\
     ~~~~~~~~~~~~:consist-of (t / target-01 \\
     ~~~~~~~~~~~~~~~~:ARG1 (a / aircraft \\
     ~~~~~~~~~~~~~~~~~~~~:ARG0-of (t2 / traffic-01 \\ 
     ~~~~~~~~~~~~~~~~~~~~~~~~:ARG1 (d / drug))))) \\
     ~~~~~~~~:time (c3 / current)) \\
     ~~~~:domain (c2 / country \\
     ~~~~~~~~:wiki ``Colombia'' \\
     ~~~~~~~~:name (n / name :op1 ``Colombia''))) \\
     \textbf{Ref}: colombia is the only country that currently has a policy of targeting drug trafficking aircraft .\\
     \textbf{Baseline}: colombia is the only country \textcolor{red}{with drug trafficking policy} .\\
     \textbf{Our approach}: colombia is the only country with the current \textcolor{blue}{policy of targets for drug trafficking aircraft} . \\
     \bottomrule
    \end{tabularx}
    \caption{Example system outputs.}
    \label{tab:case_study}
    \vspace{-1.0em}
\end{table}

For the first example, the baseline output drops the key predicate ``\emph{recommend}'' and fails to preserve the fact that ``\emph{you}'' is the subject of ``\emph{go}''.
The reason can be that ``\emph{I should go to}'' 
occurs frequently
in the training corpus.
On the other hand, the extra signals produced by our multi-view framework enhance the input semantic information, guiding our model to generate a correct sentence with the exact meaning of the input AMR.

The second example shows a similar situation, where the baseline generates a natural yet short sentence that drops some important information from the input graph.
As a result of the information loss, the resulting sentence conveys an opposite meaning (``\emph{with drug trafficking policy}'') to the input (``\emph{targeting drug trafficking aircraft}'').
This is a typical problem suffered by many neural graph-to-sequence models.
Our multi-view framework helps recover the correct meaning: ``\emph{policy of target for drug trafficking aircraft}''.

\subsection{Ablation Study}

\begin{table}
    \centering
    \begin{tabular}{lc}
    \toprule
    Model & BLEU \\
    \midrule
    RA-Trans-F-ours + Loss 1 & 30.47 \\
    ~~~~~ w/o edge label & 29.39 \\
    \hdashline
    RA-Trans-F-ours + Loss 2 & 31.13 \\
    ~~~~~ w/o edge label & 30.36 \\
    ~~~~~ random linearization & 31.07 \\
    \bottomrule
    \end{tabular}
    \caption{Ablation study for both views.}
    \label{tab:ablation_linear}
    \vspace{-1.0em}
\end{table}

As shown in Table \ref{tab:ablation_linear}, we conduct an ablation study on LDC2015E86 to analyze how important each part of the input graphs is under our framework.
For \emph{Loss 1}, we test the situation when no edge labels are available, and as a result, we observe a large performance drop of 1.0+ BLEU points.
This is quite intuitive, because edge labels carry important relational knowledge between the two connected nodes.
Therefore, discarding these labels will cause loss of significant semantic information.

For \emph{Loss 2}, we also observe a large performance decrease when edge labels are dropped, confirming the observation for \emph{Loss 1}.
In addition, we study the effect of random graph linearization, where the order for picking children is random rather than following the left-to-right order at each stage of the depth-first traversal procedure.
The motivation is to investigate the robustness of \emph{Loss 2} regarding input variances, as an organized input order (such as an alphabetical order for children) may not be available for certain graph-to-sequence tasks.
We observe a marginal performance drop of less than 0.1 BLEU points, indicating that our approach is very robust for input variances.
It is likely because different linearization results still indicate the same graph.
Besides, one previous study \citep{konstas2017neural} shows a very similar observation.

\subsection{Main Results on LDC2017T10}
\label{sec:results_2017}

\begin{table}
    \centering
    \begin{tabular}{lcc}
    \toprule
    Model & BLEU & Meteor \\
    \midrule
    DCGCN & 27.60 & -- \\
    RA-Trans-CNN & 31.82 & 36.38 \\
    GPT-2L & 32.47 & 36.80 \\
    GPT-2L re-scoring & 33.02 & 37.68 \\
    \hline
    RA-Trans-F-ours & 31.77 & 37.2 \\
    ~~~~~+~Loss 1 & 33.98 & 37.5 \\
    ~~~~~+~Loss 2 & 34.13 & 37.8 \\
    ~~~~~+~Both & 34.21 & 38.0 \\
    \bottomrule
    \end{tabular}
    \caption{Main test results on LDC2017T10.}
    \label{tab:main_2017}
\end{table}

Table \ref{tab:main_2017} compares our results on LDC2017T10 with the highest numbers reported by single models without extra silver training data.
\emph{GPT-2L} \cite{mager-etal-2020-gpt} applies a GPT-2 Large model \cite{radford2019language} on linearized AMR graphs, and \emph{GPT-2L re-scoring} further performs re-scoring based on AMR back-parsing for post processing.
\emph{RA-Trans-CNN} is another model by \citet{zhu2019modeling} that adopt a convolutional neural network \cite{lecun1990handwritten} to model the relation path for each node pair.
Again, the \emph{RA-Trans-F} baseline achieves a comparable score with \emph{RA-Trans-CNN}, and our approach improves the baseline by nearly 2.5 BLEU points, indicating its superiority.

Regarding Meteor score, our advantage (1.62 points) over the previous state-of-the-art system on this dataset is larger than that (0.75 points) on LDC2015E86.
Since LDC2017T10 has almost one time more training instances than LDC2015E86, we may conclude that the problem of dropping input information may not be effectively reduced by simply adding more supervised data, and as a result, our approach can still be effective on a larger dataset.
This conclusion can also be confirmed by comparing the gains of our approach on both AMR datasets regarding BLEU score (2.3 vs 2.5).

\subsection{Main Results on WebNLG}

Table \ref{tab:main_webnlg} shows the comparison of our results with previous results on the WebNLG testset.
\emph{ADAPT} \citep{gardent2017webnlg} is based on the standard encoder-decoder architecture \citep{cho2014learning} with byte pair encoding \citep{sennrich2016neural}, and it was the best system of the challenge.
GCN$_{EC}$ \citep{marcheggiani2018deep} is a recent model using a graph convolution network \citep{kipf2016semi} for encoding KGs.

Our baseline shows a comparable performance with the previous state of the art.
Based on this baseline, applying either loss leads to a significant improvement, and their combination brings a gain of more than 2 BLEU points.
Although the baseline already achieves a very high BLEU score, yet the gains on this task are still comparable with those on AMR-to-text generation.
This observation may imply that the problem of missing input structural knowledge can be ubiquitous among many graph-to-text problems, and as a result, our approach can be widely helpful across many tasks.

\begin{table}
    \centering
    \begin{tabular}{lcc}
    \toprule
    Model & BLEU & Meteor \\
    \midrule
    ADAPT & 60.59 & 44.0 \\
    GCN$_{EC}$ & 55.90 & 39.0 \\
    \midrule
    RA-Trans-F-ours & 60.51 & 42.2 \\
    ~~~~~+~Loss 1 & 61.78 & 43.6 \\
    ~~~~~+~Loss 2 & 62.29 & 43.5 \\
    ~~~~~+~Both & \textbf{62.89} & 44.2 \\
    \bottomrule
    \end{tabular}
    \caption{Main test results on WebNLG}
    \label{tab:main_webnlg}
\end{table}

Following previous work, we also report Meteor scores, where
our approach shows a gain of 2 points against the baseline and our final number is comparable with \emph{ADAPT}.
Similar with the gains on the BLEU metric, \emph{Loss 1} is comparable with \emph{Loss 2} regarding Meteor, and their combination is more useful than applying each own.

\section{Conclusion}

We proposed reconstructing input graphs as autoencoding processes to encourage preserving the input semantic information for graph-to-text generation.
In particular, the auxiliary losses for recovering two complementary views (triple relations and linearized graph) of input graphs are introduced, so that our model \emph{is trained} to retain input structures for better generation.
Our training framework is general for different graph types.
Experiments on two benchmarks showed the effectiveness of our framework under both the automatic BLEU metric and human judgements.

\section*{Acknowledge}

Both Te An and Jinsong Su were supported by the National Key R\&D Program of China (No. 2019QY1803), National Natural Science Foundation of China (No. 61672440), and the Scientific Research Project of National Language Committee of China (No. YB135-49).
Yue Zhang was supported by the joint research program between BriteDreams robotics and Westlake University.
We thank the anonymous reviewers for their constructive suggestions.

\bibliography{acl2019}
\bibliographystyle{acl_natbib}

\end{document}